\documentclass{article} 
\usepackage{nips13submit_e,times}
\usepackage{hyperref}
\usepackage{url}
\usepackage{float}
\usepackage{amsmath}
\usepackage{ textcomp }
\usepackage{graphicx}

\title{Bayesian Nonparametrics in Topic Modeling: A Brief Tutorial}

\author{
Alexander Spangher \\
Columbia University, class of 2014 \\
aas2230 \\
CS6998: High Dimensional Data Analysis
}

\nipsfinalcopy 

\begin{document}

\maketitle

\begin{abstract}
Using nonparametric methods has been increasingly explored in Bayesian hierarchical modeling as a way to increase model flexibility. Although the field shows a lot of promise, inference in many models, including Hierachical Dirichlet Processes (HDP), remain prohibitively slow. One promising path forward is to exploit the submodularity inherent in Indian Buffet Process (IBP) to derive near-optimal solutions in polynomial time. In this work, I will present a brief tutorial on Bayesian nonparametric methods, especially as they are applied to topic modeling. I will show a comparison between different non-parametric models and the current state-of-the-art parametric model, Latent Dirichlet Allocation (LDA).
\end{abstract}

\section{Research Goals:}

My goals on the outset of this experiment were to:

\begin{itemize}
\item Learn more about nonparametric models
\item Investigate different non-parametric topic models and compare them to an LDA process run on a New York Times dataset with topics $k = 55$.
\item Pick higher-performing algorithms and apply recent research in Indian Buffet Process submodularity to these algorithms to assess speed increase.
\end{itemize}

I should note that my experiments were not successful. The non-parametric topic modeling algorithms that I compared to LDA had consistently worse perplexity scores. Thus, I did not go forward with proposed implementation of submodular methods. This speaks less to the potential of nonparametric models then to the need for more work in the field and the need to lower the barrier of entry--I found that unifying introductory texts were far and few between. 

Thus, I will treat this paper as the start of a tutorial I plan to build out in the next few months. While I am certainly not an expert, I approached the subject as a beginner just a few weeks ago. This paper gives me an opportunity to offer a beginner's look at non-parametrics, one that I will expand on. Although I feel that I have not contributed in a significant way research-wise, if I expand this paper over the next few months to give a better high-level introduction to nonparametric modeling, this can be an important contribution.

I have tried to make my emphasis in this paper orthogonal to my presentation. As in, I will go lightly on the topics that I went heavily into during my presentation and try to go more deeply into the background of processes.

\section{Introduction}

Bayesian hierarchical techniques present unified, flexible and consistent methods for modeling real-world problems. In this section, we will review parametric models by using the Latent Dirichlet Allocation model (LDA), presented by Blei et. al. (2003) [1], as an example. We will then introduce Bayesian Processes and show how they factor into non-parametric models.

\subsection{Bayesian Parametric Models: Latent Dirichlet Allocation (LDA)}

\begin{figure}[H]
\centerline{\includegraphics[scale=.25]{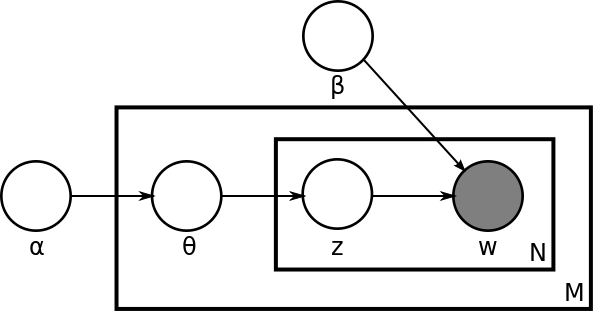}}
\caption{LDA Graphical Model [1] (Blei, 2003)}
\end{figure}

The LDA topic-model is a proto-typical generative Bayesian model. It is "generative" in the sense that it assumes that some real-world observation (in this case, word-counts), are \textit{generated} by series of draws from underlying distributions. According to Blei's hypothesis, the LDA model is generated as follows:

For each document \textbf{w} in corpus \textit{D}:
\begin{enumerate}
\item Choose $ N \sim Poisson(\kappa)$
\item Choose $\theta \sim Dir(\alpha)$
\item For each of the \textit{N} words $w_n$:
	\begin{itemize}
		\item Choose a topic $ z_n \sim Multinomial(\theta) $
		\item Choose a word $ w_n \sim Multinomial(\beta_{z_n}) $
	\end{itemize}
\end{enumerate}

These underlying distributions are hidden from us when we observe our sample set (the corpus), but through various methods we can infer their parameterization. This allows us to predict future samples, compress information, and explain existing samples in useful ways.

The LDA is a mixture model--it assumes that each word is assigned a class, or "topic", and that each document is represented by a mixture of these topics. The overall number of topics in a corpus, \textit{k}, needs to be parameterized by the user, and this is often difficult to interpret. To overcome this limitation, we introduce non-parametric models.

\subsection{Bayesian Nonparametric Models: An Introduction}

\begin{center}
   \textit{Gaussian Processes: Function Modeling}
\end{center}

We start our discussion on nonparametrics by discussing Gaussian Processes, since their construction follows the way in which processes we actually use will be constructed, yet I feel Gaussian-\textit{anything} is a naturally easier concept to grasp. Let's say $\mathcal{X}$ is distributed according to a Gaussian Process with mean measure $m(\textbf{x})$ and covariance $k(\textbf{x},\textbf{x}^{'})$:

\begin{center}
$\mathcal{X}$ \texttildelow $\mathcal{GP}(m(\textbf{x}),k(\textbf{x},\textbf{x}^{'}))$
\end{center}

Mean measure and covariance measures are simply functions that take some possibly infinite subset \textbf{x} of Euclidean (or, more generally, Hilbert space) and return a \textit{measure}. It is most helpful when uses processes in Bayesian nonparametrics to think of a "measure" as our prior belief, expressed not as a paramter but as a mapping function.

In other words, much like a function maps an input space to a quantity, measures are used to quantize finite sets. For example, lets take the \textit{mean measure}. In the Gaussian Process, $m(\textbf{x}) = E\big[f(\textbf{x})\big]$. This maps a finite (or infinite) set \textbf{x} to an expected value. In practice, when we have no information about $f(x)$, it's common to choose $m = \textbf{0}$, as a suitable measure prior. \textit{(This is reflected in Figure 2.a which shows our prior beliefs about the  $\mathcal{GP}$, where the span of the orocess is given by the gray region, with mean 0 and constant variance.)}

Over a finite subset of Euclidean space, \textbf{x}, the measure takes an expected, real value, and the Gaussian Process takes a discrete distribution: a joint Gaussian. For example, if $x_1$, $x_2$ are subsets of the real line, then $y_1$, $y_2$ are distributed as:

\begin{center}
$(y_1, y_2)$ \texttildelow $\mathcal{GP}(m(\{\textbf{x}_1,\textbf{x}_2\}),k(\{\textbf{x}_1,\textbf{x}_2\},\{\textbf{x}_1,\textbf{x}_2\}^{'})) = \mathcal{N}\big(\begin{pmatrix}\mu_1\\ \mu_2 \end{pmatrix}, \Sigma)$ 
\end{center}

This joint Gaussian property of the $\mathcal{GP}$, over an infinite subspace, can describe the set of points forming functions, shown in Figure 2. We can envision starting with a finite number of discrete subsets of the real line, represented as the blue points in the diagram, and continuing to add samples. As the number of subsets we add approaches infinity, we have encompassed the entire real line, and created a joint Gaussian who's realization forms a continuous function. This ability of the $\mathcal{GP}$ to model functions allows it to be used in kernel-based non-linear regression as a prior. (For more information, see [3])

\begin{figure}[H]
\centerline{\includegraphics[scale=.6]{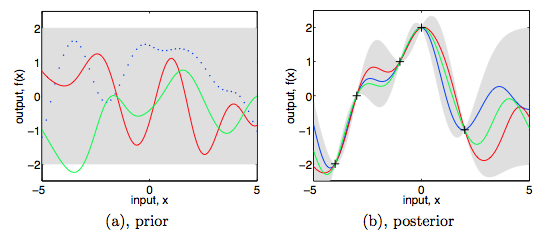}}
\caption{Gaussian Process used to describe the function space. (a) Prior, with $\mathcal{GP}(\textbf{0},\Sigma)$. The blue, dotted line represents $\mathcal{GP}$ over bounded subspaces while the red and blue lines represent the unbounded cases. (b) Posterior, updated to fit observed data.  [3] (Rasmussen, 2006)}
\end{figure}

\begin{center}
   \textit{Dirichlet Processes: Probability Modeling}
\end{center}

Similarly to how the Gaussian Process can be applied over infinite subspaces to model functions, the Dirichlet Process can model continuous probability densities. This is also rather intuitive. 

Let's start with a finite set, $A$. If $\Pi$ is a partition of $A$ such that $\bigcup _{i=1}^k \Pi_i = A$ and $T_k \cap T_l = \emptyset$ for $ k \neq l$, then:
\begin{center}
$
(G(\Pi_1),...,G(\Pi_k))$ \texttildelow $Dir(\alpha H(\Pi_1),...,\alpha H(\Pi_k))
$
\end{center}

Where $G$ is drawn from a Dirichlet Process ($\mathcal{DP}$) with base measure $H$ and concentration parameter $\alpha$:

\begin{center}
	$G$ \texttildelow  $\mathcal{DP}(\alpha,H)$   where   $E\big[G(\Pi_i)\big] = H(\Pi_i)$
\end{center}

We can think of the base measure $H$ in the Dirichlet Process the same way as we think of the mean measure in the $\mathcal{GP}$, simply as a mapping between subsets of Hilbert space to reals--in this case reals $\in [0,1]$. 

Now, using a Dirichlet process to model the space of continuous probability functions makes sense when we consider the special property of the Dirichlet Distribution that a sample from the distribution is a tuple that sums to 1--essentially a probability mass function. Thus, a Dirichlet Process over infinitely many partitions gives a continuous probability distribution much the same way that a sampling from a $\mathcal{GP}$ over infinitely many subsets of the real line gives a continous function. 

The Dirichlet Process has found many uses in modeling probability distributions. One example is the Hierarchical Dirichlet Process (HDP) by Yee Whye Teh [4]. This model uses two stacked Dirichlet Processes. The first is sampled to provide the base distribution for the second, allowing a fluid construction of point mixture-modeling known as the "Chinese Restaurant Franchise". However, the HDP has noted flaws, in particular: because the HDP draws the class proportions from a dataset-level joint distribution, it makes the assumption that the weight of a component in the entire dataset (in our application, the "corpus") is correlated with the proportion of that component being expressed within a datapoint (a "document"). Or, in other words, the probability of a datapoint exhibiting a class is correlated with the weight of that class within the datapoint. Intuitively, in the topic modeling context, we might argue that rare corpus-level topics are often expressed to a great degree within specific documents, thus indicating that the HDP is flawed.

\begin{center} 
\textit{Beta Process: Point Event Modeling}
\end{center}

The Beta Process offers us a way to perform class assignment without this correlational bias. We will see more in the next section.

Taking a step back, a draw from a Beta Distribution is a special case of a Dirichlet distribution draw over over two classes. Similarly, the Beta Process is defined over the product space $\Omega \times [0,1]$. 

A draw from a Beta Process is given as:
\begin{center}
	$B = \sum_{k=1}^\infty p_k\delta_{w_k}$
\end{center}
where $B$ \texttildelow  $BP(c, B_0)$, and $\delta_{w_k}$ can be thought of as a unit measure of $w_k$. $p$ and $w$ are defined by the Levy measure, $v$, of the Beta Process' product space, (given by Hjort (1990) [6]), is

\begin{equation*}
\nu_{BP}(dpdw) = cp^{-1}(1-p)^{c-q}dpB_0(dw)
\end{equation*}

which can be passed as the mean measure to a Poisson Process. (Where $c > 0$ is the concentration parameter and $B_0$ is the continuous-finite base measure). (This is commonly used in Levy-Khintchine formulations of stochastic processes.)

Therefore, we can see that BP draws need not sum to one. Taken over infinite subspaces, the Beta Process can be used to model CDF's, as point measurements are $\in [0,1]$, but draws are not dependent on the sum of subspaces, like the Dirichlet. Taking the aggregate of point measurements, we can derive CDF's, which are especially useful in fields like survival analysis.

As we can see, $w_j$ point measurements can also be used to model class assigments. Thus, Beta Processes can effectively model each class-membership separately. Per datapoint, probability mass behind $n$ subspaces, or classes, is no longer bounded by one as in the Dirichlet Process. (For more on Beta Processes see Paisley [7] and Zhou [8].)

\subsection{Bernouilli Processes and the Indian Buffet Process}

Here we construct such a mixed-membership modeling scheme. The Beta Process is conjugate to the Bernouilli family. (A Bernouilli process is a very simple stochastic process that can be thought of simply as a sequence of coin flips with probability \textit{p}). This makes the Beta-Bernouilli pairing an ideal candidate for mixed-membership applications like topic modeling, where each point data point is expressed as a combination of latent classes.

Griffiths and Gharamani [9] recognized this in 2005, and constructed the Indian Buffet Process (IBP) by marginalizing out the Beta process. As Thibaux and Jordan [10] would later show in 2007, given:

\begin{center}
$B $ \texttildelow $BP(c,B_0)$
$\{X_i | B\}_{1...n}$ \texttildelow $BeP(B)$
\end{center}

where BeP is the Bernouilli Process, the Beta Process can be marginalized out to give:
\[
	X_{n+1}|X_{1...n} \sim BeP \bigg(\frac{c}{c + n}B_0 + \sum_{i=1}^n \frac{m_{n,j}}{c + n} \delta_{w_j}  \bigg)
  \]
where $B_0$ is a continuous base distribution with mass $B_0(\Omega) = \gamma$, $\delta_w$ is the unit point mass at $w$, and $m_{n,j} = \sum_{i=1}^n X_i$, or the number of datapoints in $w_j$. Thus, we see through this marginalization that the class labels $j$ are independent of order--as long as the order of $w_j$ is consistent across $X_n$, the BeP construction remains the same. This realization can be used to prove exchangeability of the beta-assigned class labels, proving that these points are a De-Finetti mixture. (For more details, see [10]. De-Finetti mixtures are great because they can be realized in provably defined probability distributions.)

If we model these process as their corresponding distributions, we can derive a probability function.

$$
\pi_k | \alpha \sim Beta(\frac{\alpha}{K}, 1)$$$$
z_{ik} | \pi_k \sim Bernoulli(\pi_k)
$$
\[
P(\textbf{Z}) = \prod_{k=1}^K \int \bigg( \prod_{i=1}^N P(z_{i,k}|\pi_k) \bigg) p(\pi_k) d\pi_k
\]
\begin{figure}[H]
\centerline{\includegraphics[scale=.4]{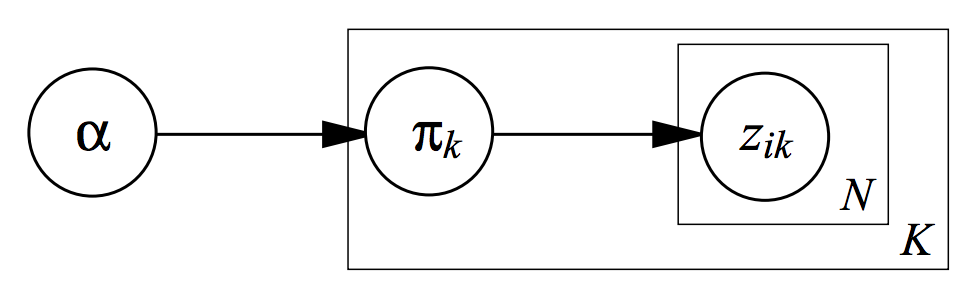}}
\caption{Graphical model for the Indian Buffet Process, [9]}
\end{figure}

Where the probability of observing a set of class assignments $\{z_{i,1},...,z_{k,i}\}$ for each datapoint $i$ is given by $P(Z)$. Under the exchangeability property of the Beta-Bernouilli construction given above, the Indian Buffet Process can be expressed as the sum of infinite classes:

\begin{align*}
P([\textbf{Z}]) &= \frac{K!}{\prod_{h=0}^{2^N-1}K_h!} P(\textbf{Z}) \\[10pt]
P([\textbf{Z}]) &= \frac{K!}{\prod_{h=0}^{2^N-1}K_h!} \prod_{k=1}^K \int \bigg( \prod_{i=1}^N P(z_{i,k}|\pi_k) \bigg) p(\pi_k) d\pi_k \\[10pt]
P([\textbf{Z}]|\alpha, \beta) &= \frac{(\alpha\beta)^{K+}}{\prod_{h=1}^{2^N-1} K_h!} e^{-a \sum_{i=1}^N\frac{\beta}{\beta + i - 1}} \prod_{k=1}^K \frac{\Gamma(m_k)\Gamma(N-m_k + \beta)}{\Gamma(N+\beta)}
\end{align*}

Where $\Gamma$ represents the Gamma function and $\alpha$ and $\beta$ are concentration parameters. $k_h$ represents the history term of the $lof$ ordering. In addition to reordering the class labels in a way that sums probability only over the "active" set $K+$ of topics, this construction collapses multiple classes assigned to the same datapoints into a single class label. For more details on the combinatorial ordering scheme see slide 22 of my presentation, included in this folder. This formulation groups the class labels of the IBP datapoints in a way that allows the probability distribution to remain well-defined even as the set of classes is unbounded. Thus, we can draw inference on the infinite parameter space through a finite set of observations.

\section{Applications of the Indian Buffet Process: Experiments}

The Indian Buffet Process has been applied to a host of mixed-model schemes. Like the Hierachical Dirichlet Process, it models class membership in a non-parametric way, thus making it appealing for use in non-parametric Bayesian models. Moreso, because it escapes the correlation that HDP introduces, the IBP has been recently studied in topic modeling applications.

One simple example is the Focused Topic Model (Blei et al., 2010) [11]. The Focused Topic Model uses an IBP Compound Dirichlet Process, which effectively decouples the correlation between high corpus-level probability mass for a topic, and high document-level probability mass. The topic weights $\theta_i$ for document $i$, now, are generated from from a $Dir(\phi \cdot \textbf{b}_i)$, where $\textbf{b}_i$ represents the binary draw from an IBP. Thus, strong corpus-level topics may be "shut-off" in the document.

Archambeau et. al. take this a step further in [12] by applying an IBP compound Dirichlet to both the document topics $\theta_i$ and the words $\beta_w$. In other words, while the Focused Topic model decouples the corpus-level strength of a topic with the probability of a document expressing a topic, it still allows words to be weighted strongly under high probability topics, encouraging distributions across words to be more uniform. Archambeau et. al. add the IBP compound Dirichlet to word probabilities as well, creating a doubly compounded model, which they call LiDA. 

Mingyuan Zhou took a different approach [13]. Instead of directly incorporating the IBP, he built a topic model around the Beta-Negative Binomial Process, a hierarchical topic model which is related to the IBP (the IBP is a Beta-Bernouilli construction) but is not binary, and thus has greater expressiveness. He presented his work at the 2014 NIPS conference.

\textit{Experiments:}

I ran trials for all three of these on New York Times dataset of 5294 documents and 5065 words, which I compared against a control of LDA with 55 topics. I chose this number has it was the current 30-day window of articles that we use for training our recommendations system. I heldout 10 percent of articles for a perplexity test. The log-perplexity of a heldout set was -7.9 for LDA on 55 topics, 2.099 for LiDA, which scaled to 66 topics, 5.6 for FTM and 5.4 for BNBP.

\section{Submodularity of the IBP}

Current inference algorithms for IBP-related processes involve either sequential Gibbs samplers or sequential variational inference, effectively creating an NP-hard problem. Recent work by Reed and Griffiths [14] examining the IBP has focused on its submodularity properties. Although I explored this in detail in my presentation, I will not go deeply into it here, as it is not relevant.

\subsubsection*{Acknowledgments}

I'd like to thank Dr. Lozano and Dr. Aravkin for a truly excellent semester, and my coworkers at the New York Times--especially Daeil Kim--for suggestions and moral support. 

\section{Conclusion}

In this paper, I've tried to give a brief tutorial on some commonly used Bayesian nonparametrics, as well as explain some motivations behind their use. I've tried to steer clear of the euphemisms many introductory reviews use to explain nonparametrics, like stick-breaking processes or poyla-urns, in favor of a more general explanations. I've reviewed some recent implementations of nonparametric models and compared them on a single dataset. 

Although the performance of the models I tested was generally lacking and the methods were slow, I still feel that nonparametrics offer a sophisticated approach towards constructing flexible Bayesian models. With more research, the I'm confident that the field could produce some promising results.

\subsubsection*{References}

\small{
[1] Blei, David M., Ng, Andrew Y. \& Jordan, M.I. (1995) Latent Dirichlet Allocation. John Lafferty (eds.), {\it Journal of Machine Learning Research} {\bf 3}, pp. 993-1022.

[2] Ibrahim, Joseph Georgy, and Ming Chen {\it Bayesian Survival Analysis.}
New York: Springer, 2001. Print.

[3] Rasmussen, Carl Edward, and Christopher K.I. Williams {\it Gaussian Processes for Machine Learning.}
Cambridge, Mass.: MIT, 2006. Print.

[4] Teh, Yee Whye, Michael I Jordan, Matthew J Beal, and David M Blei. "Hierarchical Dirichlet Processes." {\it Journal of the American Statistical Association}: 1566-581. Print.

[5] Sudderth, Eric. "Graphical Models for Visual Object Recognition and Tracking." {\it Doctoral Thesis, Massachusetts Institute of Technology (2006).}

[6] Hjort, Nils Lid. "Nonparametric Bayes Estimators Based on Beta Processes in Models for Life History Data." {\it The Annals of Statistics (1990)}: 1259-294. Print.

[7] John Paisley and Michael Jordan. "A Constructive Definition of the Beta Process". Technical Report. 2014. Print

[8] Zhou, Mingyuan, Lauren Hannah, David Dunson, and Lawrence Carin. "Beta-Negative Binomial Process and Poisson Factor Analysis." {\it Proceedings of the 15th International Conference on Artificial Intelligence and Statistics (AISTATS)} (2012). 

[9] Griffiths, Thomas, and Zoubin Ghahramani. "Infinite Latent Feature Models and the Indian Buffet Process." {\it Presented at NIPS, 2005} (2005). Print.

[10] Thibaux, Romain, and Michael Jordan. "Hierarchical Beta Processes and the Indian Buffet Process." {\it Presented at AISTATS 2007 Conference} (2007).

[11] Williamson, Sinead, Chong Wang, Katherine Heller, and David Blei. "The IBP Compound Dirichlet Process and Its Application to Focused Topic Modeling." {\it Proceedings of the 27th International Conference on Machine Learning} (2010).

[12] Archambeau, Cedric, Balaji Lakshminarayanan, and Guillaume Bouchard. "Latent IBP Compound Dirichlet Allocation." {\it IEEE Transactions on Pattern Analysis and Machine Intelligence:} \textbf{1}. Print.

[13] Zhou, Mingyuan. "Beta-Negative Binomial Process and Exchangeable Random Partitions for Mixed-Membership Modeling." {\it Presented at Neural Information Processing Systems 2014.} (2014)

[14] Reed, Colorado, and Zoubin Ghahramani. "Scaling the Indian Buffet Process via Submodular Maximization." {\it Presented at International Conference on Machine Learning 2013} (2013). 

\end{document}